\documentclass{llncs}

\usepackage{geometry}
\usepackage{color}
\usepackage{graphicx}
\usepackage{multirow}

\begin{document}

\title{Detecting the Moment of Completion:\\Temporal Models for Localising Action Completion}
\titlerunning{Completion Detection}  % abbreviated title (for running head)
\author{Farnoosh Heidarivincheh \and Majid Mirmehdi \and Dima Damen}
\authorrunning{Heidarivincheh et al.} % abbreviated author list (for running head)
%\tocauthor{Farnoosh Heidarivincheh, Majid Mirmehdi, and Dima Damen}
%
\institute{Computer Science Department,
 University of Bristol,
 Bristol, BS8 1UB, UK\\
\email{farnoosh.heidarivincheh@bristol.ac.uk}
}

\maketitle              % typeset the title of the contribution
\thispagestyle{plain}

\begin{abstract}
Action completion detection is the problem of modelling the action's progression towards localising the \textit{moment} of completion - when the action's goal is confidently considered achieved. In this work, we assess the ability of two temporal models, namely Hidden Markov Models (HMM) and Long-Short Term Memory (LSTM), to localise completion for six object interactions: \textit{switch, plug, open, pull, pick} and \textit{drink}. We use a supervised approach, where annotations of {\textit{pre-completion} and \textit{post-completion}} frames are available per action, and fine-tuned CNN features are used to train temporal models. 
Tested on the Action-Completion-2016 dataset from~\cite{Heidari}, we detect completion within 10 frames of annotations for $\sim$75\% of {completed} action sequences using both temporal models. Results show that fine-tuned CNN features outperform hand-crafted features for localisation, and that observing incomplete instances is necessary when incomplete sequences are also present in the test set.

\end{abstract}

\section{Introduction}
\label{sec:intro}
An action is defined, based on the Oxford dictionary, as \textit{the fact or process of doing something, typically to achieve an aim}.
Previous works on action recognition from visual data have overlooked assessing whether the action's \textit{aim} has actually been achieved, rather than merely attempted. 
{Similarly, action localisation approaches detect the start and the end of an action's attempt, without assessing whether the \textit{aim} has been successfully achieved.}
The notion of assessing completion was introduced in~\cite{Heidari}
{where} the performance of a variety of {hand-crafted} RGB-D features to distinguish between complete and incomplete sequences {was} reported, using a binary classifier which {was} learnt per action. 
However, the approach in~\cite{Heidari} considered the sequence as a whole and classified it as either complete or incomplete. 
In this work, we
closely examine the progression of the action and attempt to locate the \textit{moment} in time when the action can indeed be considered completed. Temporal localisation of the moment of completion is a step towards an online approach that can detect completion once the action's goal has been achieved.
Completion could prompt reminders or guidance for the next action.
For example, when preparing a food recipe, the user could be notified to wash the vegetables once these have been picked up. 

Thus, we define the problem of `{detecting the moment of completion}' as detecting the frame that separates {\textit{pre-completion} from \textit{post-completion}} per sequence, when present. 
Note that the `moment of completion' is different from the typical `start'/`end' frames in action localisation. The former focuses on the action's goal, while the latter focuses on the start and conclusion of any motion of relevance to the action. 
{Consider the action `drink'; the action is completed when the subject consumes some of the beverage from a cup or glass. Traditionally, localisation approaches locate the first movement of the hand as it lifts the cup up to the mouth, all the way to putting the cup down. Even if drinking does not actually take place, the start and end of the hand motion is localised. In contrast, the moment of completion is as soon as the subject actually drinks the beverage. The moment of completion cannot be localised when the action's goal has not been completed.}

\begin{figure}[t]
\begin{center}
\includegraphics[width=1\textwidth]{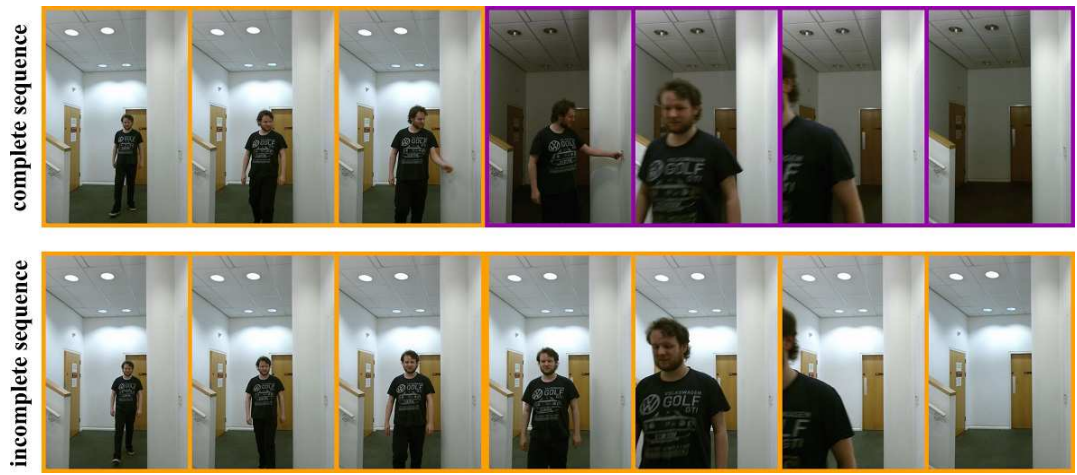}
\caption{Complete and incomplete sequences for action \textit{switch} are performed similarly up to the completion moment, {after which an} observer would confidently confirm completion. In this example, completion occurs as soon as the light is off. The {\textit{pre}- and \textit{post-completion}} frames are shown in orange and purple {bordering}, respectively.}
\label{fig:switch_example}
\end{center}
\end{figure}

In detecting the moment of completion, we take a supervised approach, where we observe complete and incomplete sequences for the same action.
Up to the moment of completion, both complete and incomplete sequences would be similar, and an observer cannot confidently say if the action is going to be completed. The sequences differ {after} the completion takes place.
Fig.~\ref{fig:switch_example} illustrates this notion using two sequences of action \textit{switch} from the dataset in~\cite{Heidari}. {In this example, our approach detects the moment the light is observed to be switched off, as the moment of completion - whereas in \cite{Heidari} the aim was to label the entire sequence as complete/incomplete}.
Completing an action makes changes in the scene which can be considered as visual cues for detecting completion. We propose {an approach} for completion detection in which these learnt signatures
are trained using Convolutional Neural Network (CNN) features along with a temporal model to encode the action's evolution, similar to~\cite{Ma2016,Yeung2015,Lei2016}.
This approach is particularly useful for action/activity detection tasks, where the classification per frame depends not only on the observations of the current frame but also on the information passed from previous frames.
Hidden Markov Models (HMMs) and Long-Short Temporal Memory models (LSTMs) are two temporal models, suitable for dealing with sequential input data and capable of handling frame by frame analysis of an action sequence. 
In this work, we compare HMM and LSTM for completion detection {on six action classes}.

The remainder of this paper is organised as follows: related work in Sec.~\ref{sec:rel_work}, the proposed method and features used in Sec.~\ref{sec:ac_model}, results in Sec.~\ref{sec:res} and conclusion and future work in Sec.~\ref{sec:Conc}.

\section{Related Work}
\label{sec:rel_work}

We now present a brief review of related works in action recognition/detection and discuss which methods could be applied for completion detection. 

Remarkable results for action recognition/detection have been achieved through extracting robust features, encoding them, then feeding them into classification methods~\cite{Kang2016}. For example, ~\cite{Zhang2014,Laptev2008,Niebles2008,Gilbert2008,Matikainen2009,Wang2013_1,Wang2013_2,Wang2016}  extract space-time features from interest points or dense trajectories, encode them using BoWs or FVs, and then use SVM to recognise the actions. Such approaches do not model the temporal dynamics of actions explicitly, thus cannot provide per-frame detection.
Recently, CNN features of spatial and temporal information have been used for action recognition/detection using dual streams~\cite{Simonyan2014,Feichtenhofer2016} or 3D convolutions~\cite{Ji2013,Tran2015}.
A seminal work of Wang \textit{et al.} \cite{Wang2016} encodes the transformations an action introduces to the scene from precondition into effect.
Although their approach is similar to ours, focussing on the changes that actions make into the scene, they perform action recognition without an explicit temporal encoding, which makes the approach unsuitable for frame level analysis or detecting the moment of completion. 

Many recent works on action recognition/detection, such as~\cite{Ma2016,Yeung2015,Lei2016,Donahue2015,Ng2015}, take advantage of CNN models to extract features from their last fully-connected layers in order to have a good representation of spatial information per frame. Then, they also {train} a temporal model on top of the extracted features to {encode} the dynamics of the action. This approach suits our completion detection task in two ways. First, extracting features from the CNN helps to encode the appearance changes in the scene. Second, it uses a temporal model and can provide per-frame decisions, while considering the context.
Two temporal models widely applied for action recognition/detection are HMMs and LSTMs.
HMM and its variants have been widely used in action recognition/detection to model temporal dynamics of actions \cite{Brand1997,Oliver2000,Ramanan2003,Weinland2007,Yu2009,Kellokumpu2011}. For this purpose, an HMM model is usually trained per desired class. Then, the likelihood of a test sequence to belong to each model is evaluated and classification is performed by specifying the model which produces the maximum likelihood for the sequence.
To detect completion, we {similarly} learn a single supervised HMM model {per action}, with two hidden states {\textit{pre-completion}/\textit{post-completion}}, 
%for all the sequences of that action and 
then use the Viterbi algorithm to find the most likely label sequence.
LSTM, as another temporal model designed to deal with sequential information, is able to capture long-term dependencies in the input data, using the memory inside each cell.
In many recent works, LSTMs have achieved promising results for the task of action recognition/detection, such as  \cite{Baccouche2010,Donahue2015,Grushin2013,Veeriah2015,Ma2016,Yeung2015}.

The closest work to ours,~\cite{Heidari}, extracts skeletal features from depth data which are temporally encoded by Fourier temporal pyramid for recognising sequence-level completion. {As they use a sequence-level representation, the method is not suitable for per-frame decisions.} 

In this work, we go beyond sequence-level completion to, (i) achieve frame-level detection of the \textit{moment} of completion, (ii) show {fine-tuned appearance features outperform pre-trained and} previously used skeletal features for completion detection, (iii) report results on {six actions from} the previously introduced action completion dataset~\cite{Heidari}, and (iv) compare per-action HMM and LSTM temporal models. Using fine-tuned appearance features, we detect the moment of completion for six different actions {- within 10 frames -} for 75.1\% and 74.6\% of the sequences by HMM and LSTM models, respectively.

\begin{figure}[t]
\begin{center}
\includegraphics[width=1\textwidth]{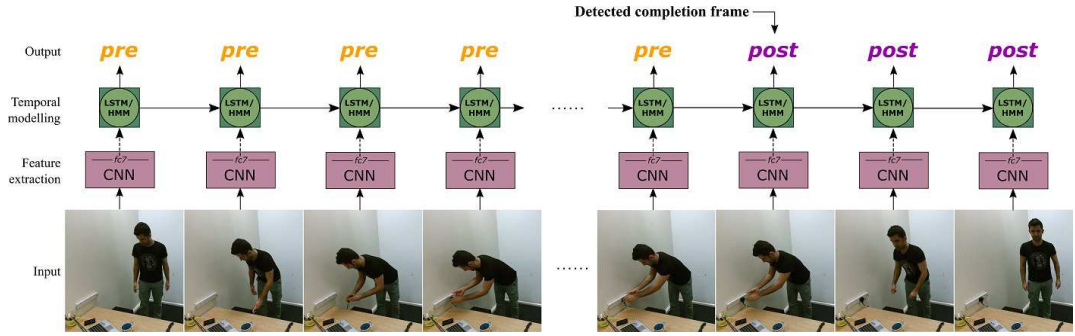}
\caption{Overall scheme of the proposed method for completion detection (of action \textit{plug}):: CNN features are extracted per frame and fed into the temporal model (LSTM or HMM). The outputs \textit{pre} and \textit{post} represent {\textit{pre-} and \textit{post-completion}} labels, respectively.}
\label{fig:model}
\end{center}
\end{figure}

\begin{figure}
\begin{center}
\includegraphics[width=0.4\textwidth]{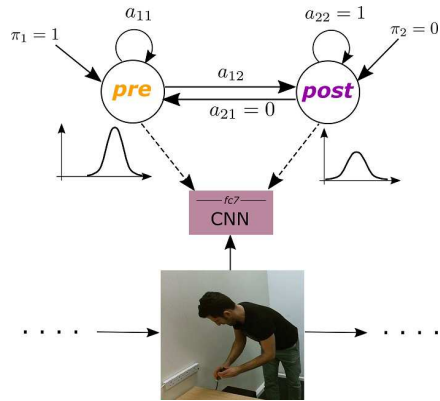}
\caption{{An HMM model for completion detection encodes that the model starts at \textit{pre-} state, and cannot transiting back to \textit{pre-} state once at \textit{post-} state.}}
\label{fig:hmm}
\end{center}
\end{figure}

\section{Proposed Method for Detecting Completion}
\label{sec:ac_model}

To present our proposal for detecting action completion, we first consider the underlying assumptions of the model, and then discuss how we use {per-frame features and train} two temporal models, HMMs and LSTMs. {We make three reasonable assumptions, namely;}
\begin{itemize}
\item \textbf{Temporally Segmented Sequences:} We assume sequences include a single instance of the action that has not been completed at the beginning of the sequence, and the action is not repeated within the sequence. Accordingly, we only need to detect a single moment of completion per sequence, if {at all}. The sequences do not include cases where the subject undertakes steps to undo the completion, e.g. a person places the cup down after picking it up.  
\item \textbf{Momentary Completion}: we assume completion could be detected using a single frame in the sequence. This is typically the first frame where a human observer would be sufficiently confident that the goal has been achieved. All frames prior to the completion localisation are {\textit{pre-completion}} and all frames {from the completion moment onwards} indicate {\textit{post-completion}}. For an incomplete sequence, all the frames would be labelled as {\textit{pre-completion}}.
\item \textbf{Uniform Prior for Completion:} in a Bayesian setting one might wish to consider the prior probability of an action being completed. Consider that the prior probability for a person to forget to switch {off} the light in their home is 0.01. A method that thus assumes completion for every sequence will have a high accuracy but is nevertheless not suitable for completion detection. Thus, we assume a uniform prior (i.e.~50-50 chance) for an action to be completed.
\end{itemize}

Fig. \ref{fig:model} illustrates the overall scheme of the proposed method, 
which {relies on} per-frame features, optimised to classify {\textit{pre-completion} and \textit{post-completion}} frames. 
{We test hand-crafted, pre-trained and fine-tuned CNN features for per-frame representations. For hand-crafted features, we evaluate the best skeletal feature from~\cite{Heidari}, namely Joint-Velocities (JV), as well as the {best} raw-depth feature, namely Local Occupancy Patterns (LOP)~\cite{Wang2012_CVPR}. We also test hand-crafted RGB features, namely Histogram of Oriented Gradients (HOG). For CNN features, we use the ``very deep VGG-16-layer'' model~\cite{Simonyan2014_CoRR} trained on UCF101 action dataset~\cite{Simonyan2014}, as pre-trained features.
We also fine-tune these models per action for two classes: {\textit{pre-completion} and \textit{post-completion}}, using leave-one-subject-out.
Features are then extracted from the \textit{fc7} layer for both the pre-trained and fine-tuned CNNs. 
}

Completion detection is then trained as a temporal model {and evaluated using two temporal approaches}:

\noindent \textbf{HMM:} A supervised HMM model with two hidden states is trained per action. Fig.~\ref{fig:hmm} illustrates this model. The parameters of this {HMM}, i.e. initial and transition probabilities, in addition to covariance matrices and mean vectors of two multivariate Gaussians (for the two states of {\textit{pre-} and \textit{post-completion}}) are computed from the training data.
Note that, due to the assumptions made for completion detection, in this supervised HMM model, the initial probabilities for {\textit{pre-} and \textit{post- completion}} states are learned from training data to be 1 and 0, respectively.
Also, the probability of transitioning from {\textit{post-completion} state to \textit{pre-completion}} state is learned to be 0, {as the sequence remains in {\textit{post-completion}} state once it transitions to it.

\noindent \textbf{LSTM:} The LSTM model is also trained per action. For this purpose, {the frames' CNN features}, along with the frames' labels {(\textit{pre-} or \textit{post-completion})}, are fed into an LSTM network. {The LSTM is optimised using Stochastic Gradient Descent with back-propagation. Also, the labels per frame are predicted using a \textit{softmax} layer on top of the LSTM hidden layer.} The details of the LSTM architecture are presented in Sec~\ref{sec:res}.

{In presenting results, we also evaluate the need for training using incomplete sequences per action. For these results, we only train on labeled complete sequences. This aims to assess when completion detection requires, or benefits from, observing incompletion.}

\section{Experimental Results}
\label{sec:res}
\noindent \textbf{Dataset and Completion Annotation:}
{We use} the RGBD-AC dataset from~\cite{Heidari},
which contains both complete and incomplete sequences for the same action. It contains 414 complete and incomplete sequences of six actions: \textit{switch, plug, open, pull, pick} and \textit{drink}.
However, the annotations provided by the dataset are sequence-level annotations. Thus, we re-annotate each sequence in the dataset for the moment of completion.
An annotation corresponding to the completion moment is assigned to every complete sequence by one human observer.

\noindent \textbf{Implementation Details:} As noted earlier, we use a pre-trained {spatial stream CNN from~\cite{Simonyan2014} which uses the VGG-16 architecture~\cite{Simonyan2014_CoRR}, fine-tuned} on UCF101 action dataset~\cite{UCF101}. We perform Leave-One-Person-Out cross validation on the 8 subjects in the dataset.
For fine-tuning, 15 epochs are performed, and the learning rate is started {at $10^{-3}$, divided by 10 {at epochs 3 and 5}, and fixed thereafter}. 
All the other hyper-parameters are set similar to~\cite{Feichtenhofer2016}. 
The 4096-dimension $fc7$ features are extracted per frame.

We test the full dimension as well as applying dimensionality reduction via PCA, as this proves essential for training the HMM.
Without applying PCA, the HMM {remains in the {\textit{pre-completion}} state and never goes to {\textit{post-completion}}. 
We keep the number of components that encapsulates at least 90\% of the variance.
LSTM however, does not require dimensionality reduction, and indeed we {see a slight drop} in performance when PCA is applied. Hence, for all results reported below, we use PCA-HMM or LSTM for the temporal models.
The LSTM model has one hidden layer with 128 units. After a random initialisation, 25 epochs are performed for training. The learning rate is $10^{-3}$ for the first epoch and is fixed at $10^{-4}$ for the next epochs.
For each test sequence, the output of HMM's Viterbi algorithm and LSTM are validated against ground truth labels.

\begin{table}[t]
\small
\begin{center}
\begin{tabular}{|c|c|c|c|c|c|c|}
\cline{3-7}
%\multicolumn{1}{c|}{ } & \multicolumn{2}{c||}{\textbf{complete}} & \multicolumn{2}{c||}{\textbf{incomplete}} & \multicolumn{2}{c|}{\textbf{total}}\\\cline{2-7}

\multicolumn{2}{c|}{ } & \textbf{JV} & \textbf{LOP} & \textbf{HOG} & \textbf{PT-CNN} & \textbf{FT-CNN} \\ \hline

\multirow{3}{*}{\textit{switch}} & precision & 78.0 & 99.0 & 99.8 & 99.7 & 99.8 \\ \cline{2-7}
& recall & 86.0 & 36.8 & 95.1 & 100 & 100 \\ \cline{2-7}
& $F_1$-score & 81.8 & 53.7 & 97.4 & \textbf{99.9} & \textbf{99.9} \\ \hline \hline
\multirow{3}{*}{\textit{plug}} & precision & 43.0 & 23.8 & 100 & 82.4 & 99.5 \\ \cline{2-7}
& recall & 76.4 & 56.3 & 9.3 & 21.0 & 94.5 \\ \cline{2-7}
& $F_1$-score & 55.0 & 33.4 & 17.1 & 33.4 & \textbf{97.0} \\ \hline \hline
\multirow{3}{*}{\textit{open}} & precision & 56.9 & NA & NA & 80.9 & 99.9 \\ \cline{2-7}
& recall & 85.1 & 0 & 0 & 9.2 & 36.2 \\ \cline{2-7}
& $F_1$-score & \textbf{68.2} & NA & NA & 16.6 & 53.2 \\ \hline \hline
\multirow{3}{*}{\textit{pull}} & precision & 57.6 & 77.8 & 100 & 100 & 99.9 \\ \cline{2-7}
& recall & 69.6 & 11.5 & 44.1 & 14.5 & 94.6 \\ \cline{2-7}
& $F_1$-score & 63.0 & 20.0 & 61.2 & 25.3 & \textbf{97.2} \\ \hline \hline
\multirow{3}{*}{\textit{pick}} & precision & 62.9 & 93.7 & 100 & 83.1 & 88.7 \\ \cline{2-7}
& recall & 82.3 & 65.7 & 3.8 & 56.3 & 97.4 \\ \cline{2-7}
& $F_1$-score & 71.3 & 77.3 & 7.3 & 67.1 & \textbf{92.9} \\ \hline \hline
\multirow{3}{*}{\textit{drink}} & precision & 66.2 & 96.1 & 95.6 & 68.2 & 84.7 \\ \cline{2-7}
& recall & 94.9 & 43.4 & 13.7 & 32.4 & 90.8 \\ \cline{2-7}
& $F_1$-score & 78.0 & 59.8 & 24.0 & 43.9 & \textbf{87.6} \\ \hline \hline \hline
\multirow{3}{*}{\textbf{Total}} & precision & 61.5 & 63.2 & 99.1 & 84.3 & 92.4 \\ \cline{2-7}
& recall & 85.1 & 36.1 & 30.2 & 39.8 & 89.0 \\ \cline{2-7}
& $F_1$-score & 71.4 & 46.0 & 46.3 & 54.1 & \textbf{90.6} \\ \hline

\end{tabular}
\end{center}
\caption{Comparing features for completion detection using PCA-HMM.}
\label{tab:ftr}
\end{table}

\noindent \textbf{Evaluation Metrics:} Due to the imbalance in the number of {\textit{pre-} and \textit{post-completion}} labels, we present precision and recall results, where the {\textit{post-completion}} is considered as the positive class. {For comparison,} $F_1$-score percentage is also reported.

While precision, recall and $F_1$-score are suitable for comparing per-frame {decisions}, we wish to evaluate the accuracy with which the moment of completion is detected.
Thus, we introduce a {cumulative} metric for evaluating completion detection that captures the difference between the labelled and estimated moment of completion, 
\begin{equation}
C(i) = \frac{1}{N} \sum_{seq = 1}^N \bigl[(p_{seq} - g_{seq})\le i \bigr] %\quad ; \quad -5s \le i \le 5s
\label{eq:res_all}
\end{equation}
\noindent where $g_{seq}$ and $p_{seq}$ correspond to the first {\textit{post-completion}} label in ground truth and predicted labels, respectively, $i$ is the shift between the predicted and true completion frames, $[.]$ is a boolean function, and $N$ is the number of sequences. We do not use the absolute difference in frames, so that for complete sequences, we can capture whether detection takes place prematurely, or is overdue per action. When completion is not detected, $p_{seq} = \infty$. Also, for incomplete sequences $g_{seq} = 0$. 

{We report these results as a cumulative graph of $i$ vs $C(i)$, for complete and incomplete sequences of each action, {independently}. The optimal plot for complete sequences is one that shows a step function from 0 to 1 when i = 0. The optimal graph for incomplete sequences would thus be a flat line at 0.} 

\noindent \textbf{Feature evaluation:}
We compare fine-tuned CNN features (FT-CNN) features to the other hand-crafted and pre-trained features using PCA-HMM in Table~\ref{tab:ftr}. As can be seen, FT-CNN has the highest $F_1$-score in total which shows its ability to represent the subtle changes in the scene observed after completion.

\noindent \textbf{Temporal Model evaluation:}
The results of comparing PCA-HMM with LSTM per action (using CNN features), are summarised in Table~\ref{tab:pca} and shown in Fig.~\ref{fig:pca}. Also, the {combined} results for all six actions in the dataset are shown in Fig.~\ref{fig:res_all}. {The figure shows that while LSTM detects completion more frequently - in fact completion is detected for 96.7\% of the complete sequences using LSTM, compared to 88.5\% using PCA-HMM,} 
LSTM is more {likely to detect completion in incomplete sequences as well}.
The {slightly delayed} detection by PCA-HMM could be because of its low transition probability (less than 1\%) to go from {\textit{pre-} to \textit{post-completion}} state.
Also, it should be noted that, since LSTM does not learn any definite constraint on its transitions, it can transit back from {\textit{post-completion} to \textit{pre-completion}}, while PCA-HMM {is forced} to remain in {\textit{post-completion}}. 
The results also show that action \textit{switch} is the nearest to the ground truth for both temporal models, which is due to the obvious change it introduces to the scene. In contrast, action \textit{open} has the {worst performance, probably due to the subtle change it introduces to the environment.}

\begin{table} 
\small
\begin{center}
\begin{tabular}{|c|c|c|c|c|c|c|c|}
\cline{3-4}
%\multicolumn{1}{c|}{ } & \multicolumn{2}{c||}{\textbf{complete}} & \multicolumn{2}{c||}{\textbf{incomplete}} & \multicolumn{2}{c|}{\textbf{total}}\\\cline{2-7}

\multicolumn{2}{c|}{ } & \textbf{LSTM} & \textbf{PCA+HMM} \\ \hline

\multirow{3}{*}{\textit{switch}} & precision & 100 & 99.8 \\ \cline{2-4}
& recall & 99.7 & 100 \\ \cline{2-4}
& $F_1$-score & \textbf{99.9} & \textbf{99.9} \\ \hline \hline

\multirow{3}{*}{\textit{plug}} & precision & 92.2 & 99.5 \\ \cline{2-4}
& recall & 96.0 & 94.5 \\ \cline{2-4}
& $F_1$-score & 94.0 & \textbf{97.0} \\ \hline \hline

\multirow{3}{*}{\textit{open}} & precision & 84.8 & 99.9 \\ \cline{2-4}
& recall & 69.1 & 36.2 \\ \cline{2-4}
& $F_1$-score & \textbf{76.2} & 53.2 \\ \hline \hline

\multirow{3}{*}{\textit{pull}} & precision & 96.0 & 99.9 \\ \cline{2-4}
& recall & 96.5 & 94.6 \\ \cline{2-4}
& $F_1$-score & 96.2 & \textbf{97.2} \\ \hline \hline

\multirow{3}{*}{\textit{pick}} & precision & 75.6 & 88.7 \\ \cline{2-4}
& recall & 96.1 & 97.4 \\ \cline{2-4}
& $F_1$-score & 84.7 & \textbf{92.9} \\ \hline \hline

\multirow{3}{*}{\textit{drink}} & precision & 82.1 & 84.7 \\ \cline{2-4}
& recall & 91.9 & 90.8 \\ \cline{2-4}
& $F_1$-score & 86.7 & \textbf{87.6} \\ \hline \hline \hline

\multirow{3}{*}{\textbf{total}} & precision & 87.7 & 92.4 \\ \cline{2-4}
& recall & 92.7 & 89.0 \\ \cline{2-4}
& $F_1$-score & 90.1 & \textbf{90.6} \\ \hline

\end{tabular}
\end{center}
\caption{Comparing LSTM and PCA-HMM using FT-CNN features.}
\label{tab:pca}
\end{table}

\begin{figure}
\begin{center}
\includegraphics[width=1\textwidth]{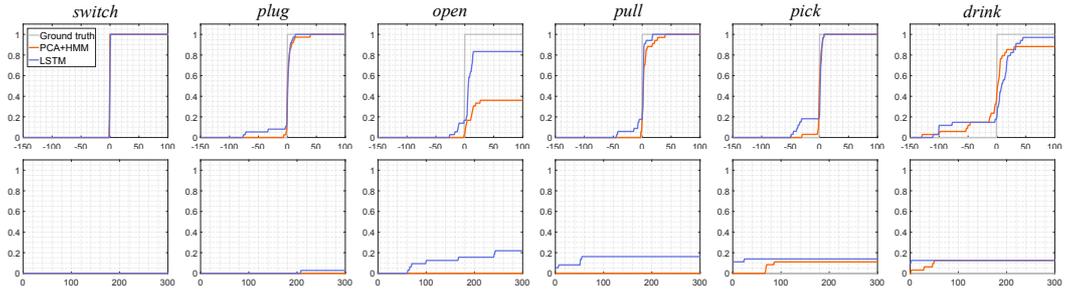}
\caption{LSTM vs PCA-HMM {for complete (top) and incomplete (bottom)} sequences.}
\label{fig:pca}
\end{center}
\end{figure}

\begin{figure}
\begin{center}
\includegraphics[width=0.4\textwidth]{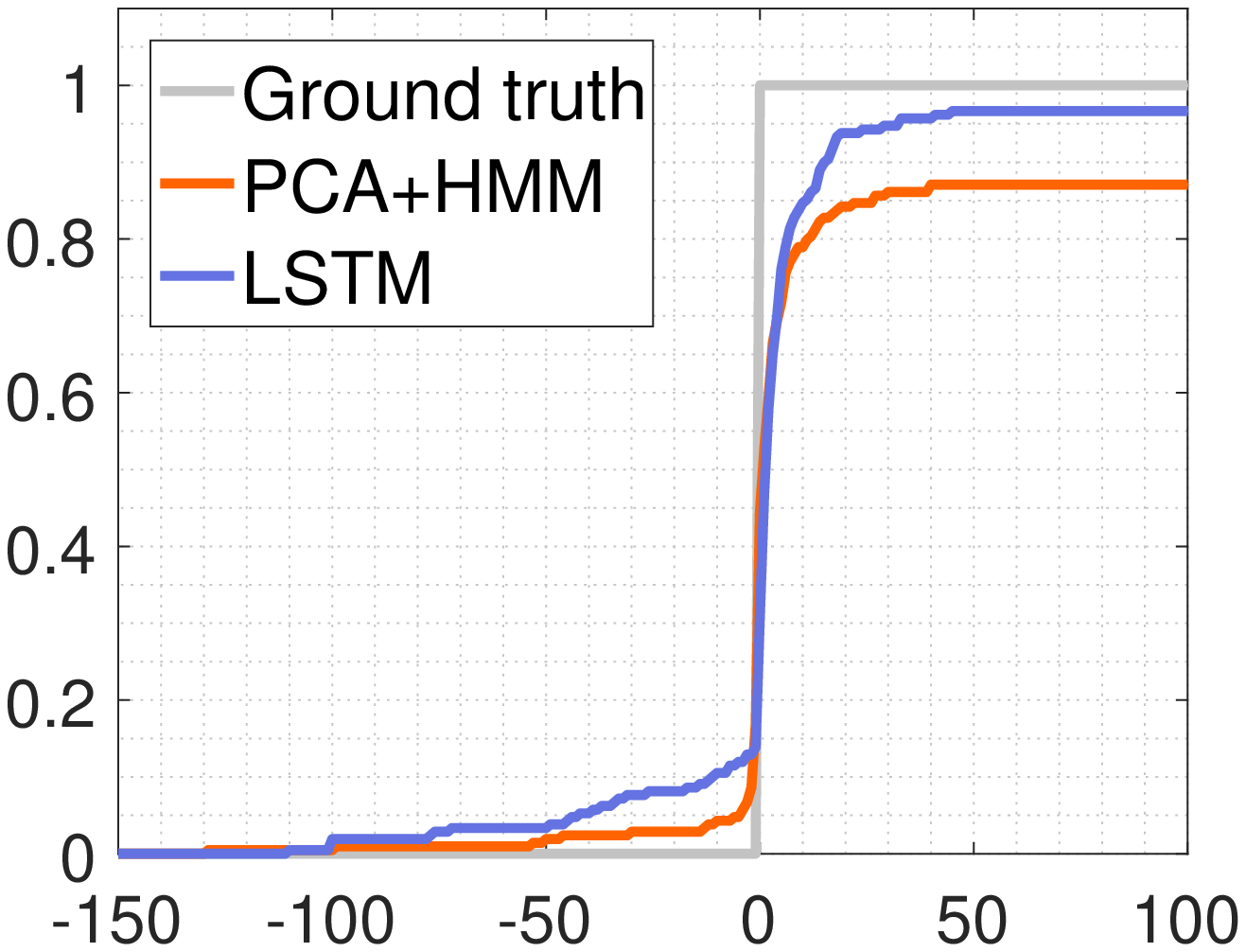}\hspace{30pt}
\includegraphics[width=0.4\textwidth]{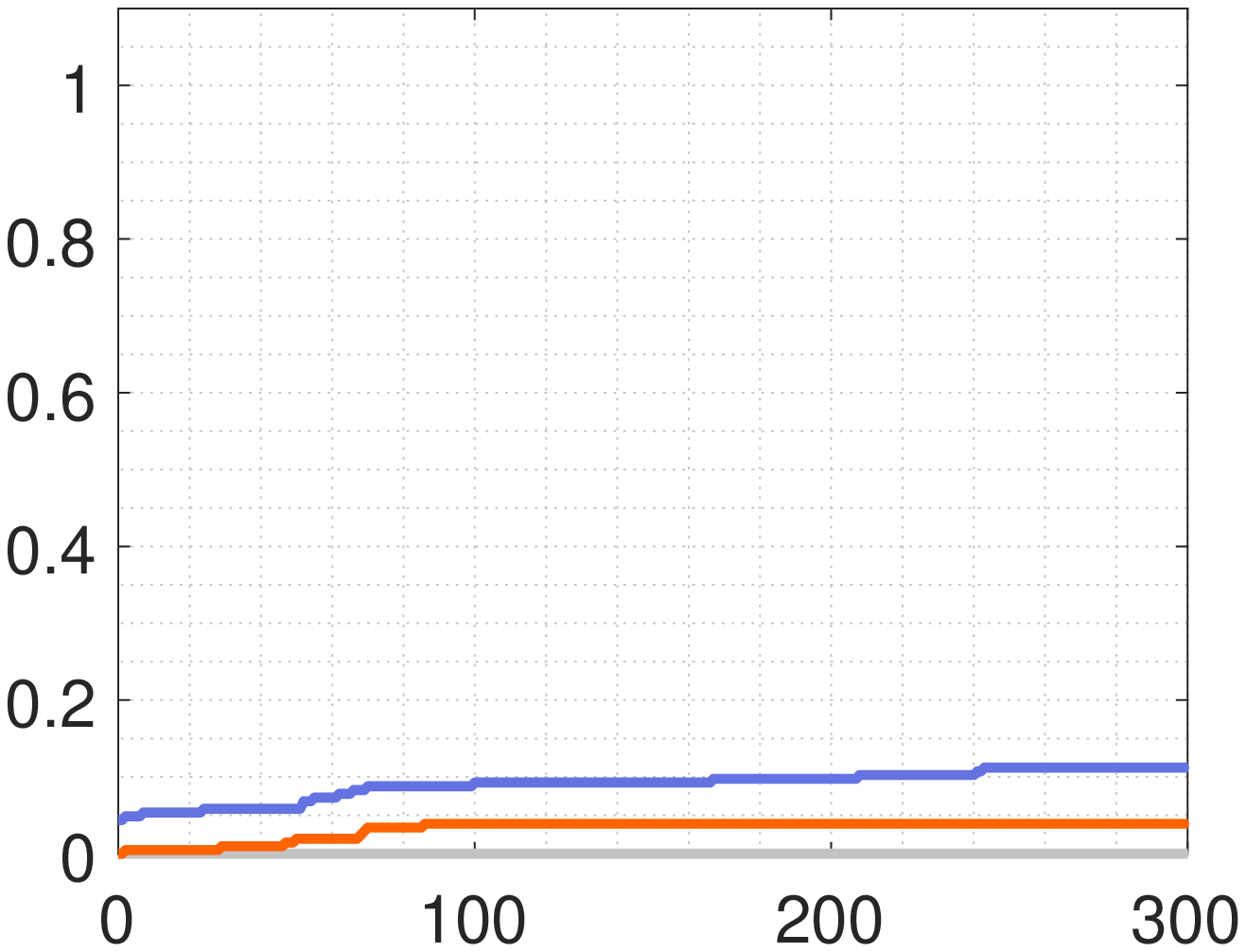}
\caption{Results for complete (left) and incomplete (right) sequences across the six actions}
\label{fig:res_all}
\end{center}
\end{figure}

\noindent \textbf{The need for observing incomplete sequences:} In theory, completion detection could use any dataset (which only contains complete sequences), as both {\textit{pre-} and \textit{post-completion}} labels are present in complete sequences.
In this experiment, we train PCA-HMM on only complete sequences, {but test on} both complete and incomplete sequences. The results are shown in Fig~\ref{fig:complete}. As seen, {while performance is slightly improved for detecting completion in complete sequences, particularly for actions \textit{open} and \textit{pull}, the performance on incomplete sequences clearly deteriorates. The method tends to label completion when the action is not completed,}
(especially for actions \textit{switch} and \textit{drink}). This shows that training only on complete sequences may cause the CNN learn {incorrect} features for completion detection and supports the need for learning from both complete and incomplete sequences.

In Fig.~\ref{fig:res_seq}, we present some qualitative success and failure results for detecting completion.

\begin{figure}
\begin{center}
\includegraphics[width=1\textwidth]{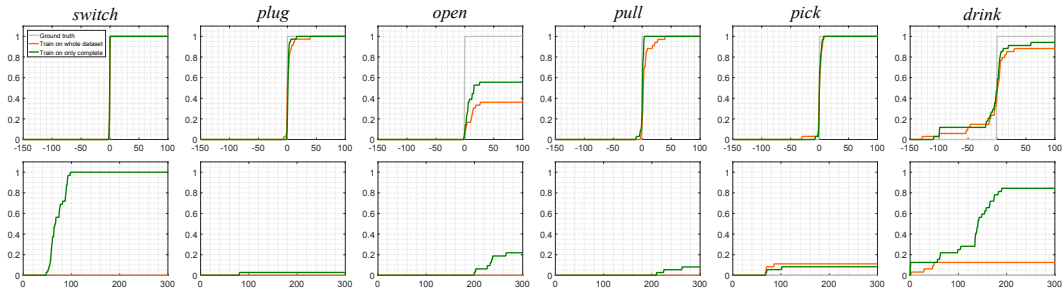}

\caption{The PCA-HMM model is trained only on complete sequences, then tested on complete (top) and incomplete (bottom) sequences.}
\label{fig:complete}
\end{center}
\end{figure}

\begin{figure}[t]
\begin{center}
\begin{minipage}{0.04\textwidth}
%\vspace*{6pt}

(a)

\vspace*{58pt}

(b)

\vspace*{58pt}

(c)

\vspace*{58pt}

(d)

\end{minipage}
%\hspace*{-15pt}
\begin{minipage}{0.9\textwidth}
 \includegraphics[width=1.0\textwidth]{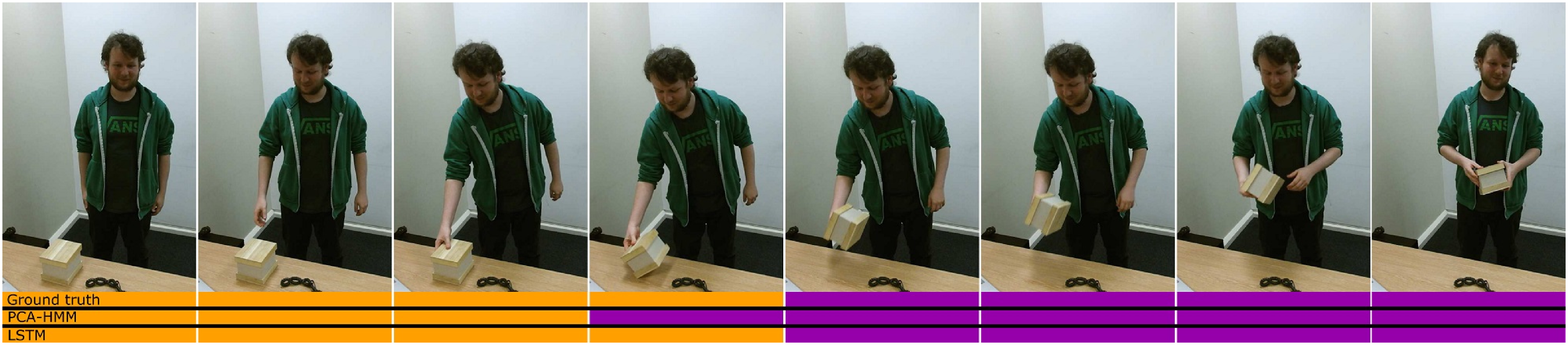}\\
\includegraphics[width=1.0\textwidth]{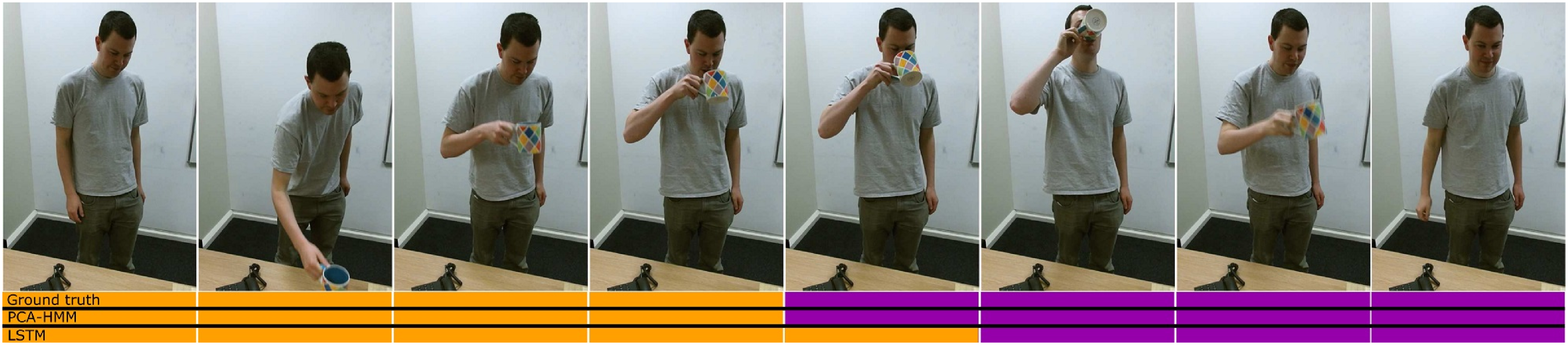}\\
\includegraphics[width=1.0\textwidth]{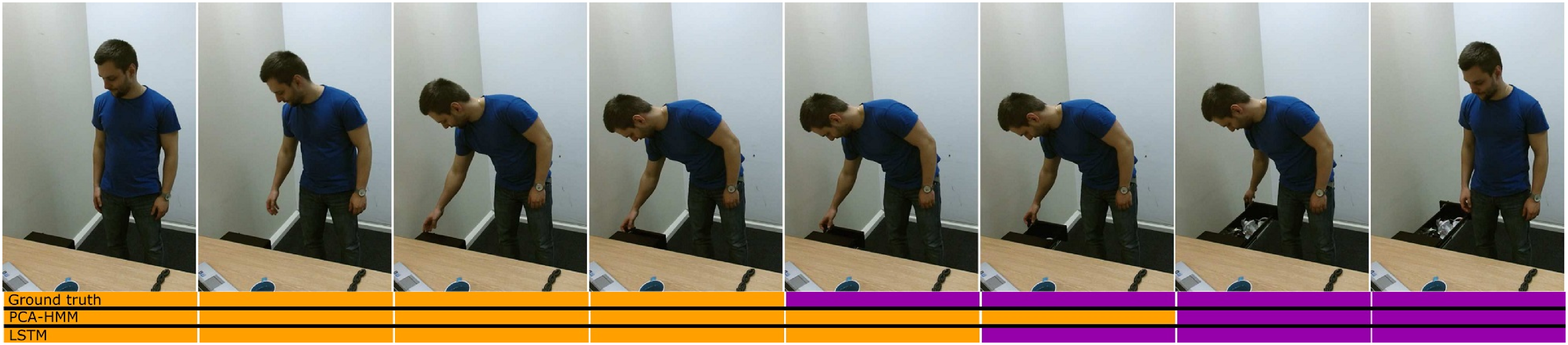}\\
\includegraphics[width=1.0\textwidth]{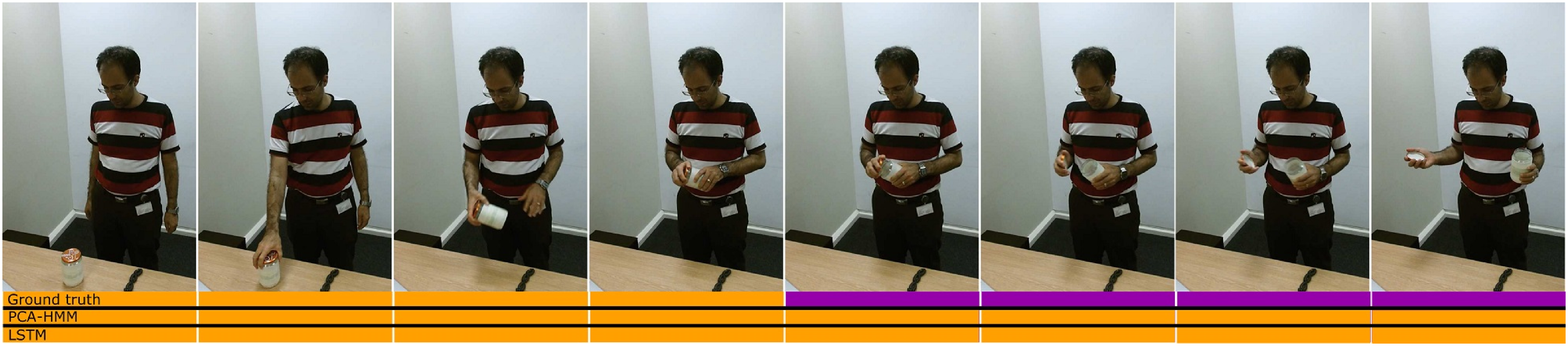}

\end{minipage}
\caption{sample sequences of success and failure for completion detection {for both PCA-HMM and LSTM: (a) PCA-HMM has an early detection by 1 frame, (b) LSTM has a late detection by 1 frame, (c) PCA-HMM is delayed by 7 frames (6 frames omitted for space), while LSTM is delayed by 1 frame, (d) both PCA-HMM and LSTM fail to detect completion.}}
\label{fig:res_seq}
\end{center}
\end{figure}

\section{Conclusion and Future Work}
\label{sec:Conc}
{This paper presents \textit{action completion detection} as the task of localising the moment in time when a human observer believes an action's goal has been achieved, for example an object has been picked up, or the light has been switched off. The approach goes beyond sequence-level recognition of completion towards frame level fine-grained perception of completion.}

{We proposed a supervised approach for detecting completion per action. Fine-tuned CNN features were extracted per-frame and used to train temporal models.} Two widely used temporal models, HMM and LSTM, were applied {and their results} were validated against the ground truth {\textit{pre-} and \textit{post-completion}} labels. We showed that, for detecting completion, {fine-tuned} CNN features outperform {pre-trained and hand-crafted} features, as they capture the subtle changes in the scene {when the action is completed}. We also introduced a metric for completion detection to {assess when completion is prematurely detected or when the decision is overdue. We showed that both temporal models (LSTM and PCA-HMM) can detect completion within 10 frames in 
$\sim$75\% of complete sequences. However, we found that PCA-HMM outperforms LSTM on incomplete sequences.}

{We aim to pursue two directions for future work. First, we aim to search for novel temporal models that particularly aim at completion detection, and can work in untrimmed videos. Second, we aim to test the method on other datasets including a variety of actions for a wider understanding of action completion detection.}

\bibliographystyle{splncs}
\bibliography{main}

\end{document}